\documentclass[conference]{IEEEtran}
\IEEEoverridecommandlockouts

\usepackage{cite}
\usepackage{amsmath,amssymb,amsthm,amsfonts}
\usepackage{graphicx}
\usepackage{textcomp}
\usepackage{xcolor}
\usepackage{xspace}
\usepackage{adjustbox} 
\usepackage{mathrsfs}%
\usepackage{textcomp}%

\usepackage{stfloats}
\usepackage{url}

\usepackage{booktabs}
\usepackage{multirow}
\usepackage{bbm}
\usepackage{thmtools}		
\usepackage{mleftright}
\usepackage{thm-restate}
\usepackage[mathic=true]{mathtools}
\usepackage{fixmath} 

\usepackage[hidelinks]{hyperref}

\usepackage{enumitem}
\usepackage{url}
\usepackage{subcaption}
\usepackage{cleveref}
\usepackage{multicol}

\usepackage{balance}

\usepackage{pifont}  %
\newcommand{\cmark}{\ding{51}} %
\newcommand{\xmark}{\ding{55}} %

\usepackage[ruled,vlined,algo2e]{algorithm2e}
\SetAlgoLined
\LinesNumbered
\SetKwInput{KwIn}{Input}
\SetKwInput{KwOut}{Output}
\SetKw{KwTo}{to}
\DontPrintSemicolon
\SetKwComment{note}{$\triangleright$ }{}
\SetFuncSty{textsc}
\SetCommentSty{textit}
\SetDataSty{texttt}
\SetKwInput{Initial}{Initial}
\SetKwInput{Parameter}{Param.}
\ResetInOut{Result} 
\let\oldnl\nl%
\newcommand{\nonl}{\renewcommand{\nl}{\let\nl\oldnl}}%
\SetKw{KwAnd}{and} %
\SetKw{KwOr}{or}
\SetKw{KwXor}{xor}
\SetKw{KwNot}{not}
\SetKw{KwTo}{to}
\SetKw{Parallel}{parallel}
\SetKw{Return}{return}

\newcommand{\NP}{\ensuremath{\mathbf{NP}}\xspace}

\newcommand{\hypergwl}{{FALCON}\xspace}
\newcommand{\hypergwlc}{{FALCON(c)}\xspace}
\newcommand{\hypergwlnc}{{FALCON(nc)}\xspace}

\newcommand{\Simplex}[1]{\Delta_{#1}}

\theoremstyle{definition}

\crefname{observation}{observation}{observations}
\Crefname{observation}{Observation}{Observations}

\crefname{assumption}{assumption}{assumptions}
\Crefname{assumption}{Assumption}{Assumptions}

\begin{document}

\title{Unsupervised Multi-Scale Gromov-Wasserstein\\Hypergraph Alignment}
\author{\IEEEauthorblockN{Lutz Oettershagen}
	\IEEEauthorblockA{\textit{University of Liverpool}\\
		Liverpool, UK \\
		\url{lutz.oettershagen@liverpool.ac.uk}}
	\and
	\IEEEauthorblockN{Honglian Wang}
	\IEEEauthorblockA{\textit{KTH Royal Institute of Technology}\\
        \textit{Digital Futures}\\
		Stockholm, Sweden \\
		\url{honglian@kth.se}}
	\and
	\IEEEauthorblockN{Aristides Gionis}
	\IEEEauthorblockA{\textit{KTH Royal Institute of Technology}\\
    \textit{Digital Futures}\\
		Stockholm, Sweden \\
		\url{argioni@kth.se}}
}

\maketitle
\begin{abstract} We study unsupervised hypergraph alignment, where the goal is to infer node correspondences between two hypergraphs using only structural information, without node features, labels, seed matches, or side information. Direct higher-order formulations can represent hyperedge interactions faithfully, but they can be computationally demanding and cumbersome for non-uniform hypergraphs. Graph-reduction approaches introduce a different challenge: clique expansions keep the alignment problem on the original node set but collapse all hyperedge evidence into one pairwise graph, whereas bipartite expansions preserve incidence structure but enlarge the problem from nodes to nodes plus hyperedges. We introduce \hypergwl{} (\textbf{F}iltration-based hypergr\textbf{A}ph a\textbf{L}ignment via \textbf{C}ross-scale \textbf{O}ptimal tra\textbf{N}sport), an unsupervised optimal-transport framework for hypergraph alignment. Instead of representing each hypergraph by a single collapsed clique graph, \hypergwl{} constructs a filtration-induced sequence of clique-based co-occurrence dissimilarity matrices and jointly aligns all levels through one shared multi-scale Gromov--Wasserstein (GW) objective. The shared transport plan enforces a globally consistent node correspondence across filtration levels while avoiding the auxiliary hyperedge nodes introduced by bipartite expansion. Experiments on perturbation benchmarks derived from real-world hypergraphs show that \hypergwl{} is robust to structural noise and in almost all cases outperforms strong graph- and hypergraph-alignment baselines. 
\end{abstract}

\section{Introduction}

Graph alignment seeks to identify a correspondence between the nodes of two graphs so that structural relationships are preserved.
The problem is \NP-hard and closely related to the Quadratic Assignment Problem (QAP)~\cite{lawler1963quadratic}, making the development of scalable and accurate algorithms particularly challenging~\cite{conte2004thirty,foggia2014graph,yan2016short,tang2025network,trung2020comparative}.
Nonetheless, graph alignment remains a core task in data mining, with applications in image processing, pattern recognition, social network analysis, and bioinformatics~\cite{bunke2000recent,sun2020survey,haller2022comparative,conte2004thirty,foggia2014graph,yan2016short}.
Many real-world systems, however, are not naturally pairwise.
Biological interactions, group communication, co-authorship, co-purchase behavior, and clinical co-occurrence data often involve relationships among sets of entities rather than pairs.
Hypergraphs provide a natural representation for such higher-order interactions by allowing each hyperedge to connect an arbitrary number of nodes~\cite{kim2024survey,lee2024survey}.\looseness=-1

\begin{figure}
	\centering
	\includegraphics[width=0.8\linewidth]{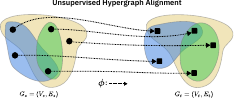}
	\caption{Given two hypergraphs $G_s$ and $G_t$, the goal is to recover a node correspondence $\phi : V_s \to V_t$ that matches an unknown hidden ground-truth mapping $\tau$ using structure alone; no node features or partial mappings are available.}
	\label{fig:intro}
\end{figure}

In this work, we study \emph{unsupervised hypergraph alignment}.
Given two related hypergraphs whose node sets are linked by an unknown latent correspondence, the goal is to recover this correspondence using only the observed hypergraph structures, without node features, hyperedge features, labels, side information, or seed matches~\cite{do2024unsupervised,ibrahim2025elruhna}.
This setting is challenging because all alignment evidence must come from higher-order structure alone.
\Cref{fig:intro} illustrates the basic alignment task.

A central obstacle in hypergraph alignment is how to represent higher-order structure.
One route is to formulate the problem directly over hyperedges, for example through tensor or other higher-order matching objectives.
Such formulations can preserve hyperedge interactions more explicitly than pairwise reductions, but they can be computationally demanding and become cumbersome for non-uniform hypergraphs, where hyperedges may have different cardinalities.
A common alternative is to reduce the hypergraph to an ordinary graph.
This reduction introduces a representation tradeoff: clique expansions keep the alignment problem on the original node set but collapse all hyperedge evidence into one pairwise object, while bipartite expansions preserve incidence structure but increase the alignment problem from $|V|$ to $|V|+|E|$ nodes~\cite{lee2024survey,zhou2006learning}.\looseness=-1

Our goal is to improve the original-node reduction setting rather than to replace all higher-order formulations.
We retain the compatibility of clique-based pairwise costs with standard optimal-transport alignment, but avoid representing the hypergraph by a single collapsed clique graph.
Instead, we use hypergraph filtration to construct a sequence of co-occurrence views.
A filtration orders hyperedges by a structural score and reveals their induced pairwise evidence progressively across levels.
Thus, \hypergwl{} addresses the loss of scale information that occurs when a hypergraph is compressed into one clique representation, while avoiding the blowup due to auxiliary hyperedge nodes introduced by bipartite expansion.

We introduce \hypergwl{} (\textbf{F}iltration-based hypergr\textbf{A}ph a\textbf{L}ignment via \textbf{C}ross-scale \textbf{O}ptimal tra\textbf{N}sport), an unsupervised framework for hypergraph alignment.
For each filtration level, \hypergwl{} builds a clique-induced node dissimilarity matrix from the active hyperedges.
Rather than aligning the levels independently, it optimizes a single shared Gromov--Wasserstein (GW) transport plan across all filtration levels.
The resulting coupling must therefore explain the source and target structures consistently across hyperedge scales.

This design combines three desirable properties within a pairwise optimal-transport framework: it keeps the alignment on the original node set, preserves scale-separated co-occurrence information that is lost in a one-shot clique representation, and produces one globally consistent node correspondence across all induced views.

We evaluate \hypergwl{} on controlled perturbation benchmarks derived from real-world hypergraphs.
The results show that \hypergwl{} is robust to structural perturbations and generally matches or outperforms strong graph- and hypergraph-alignment baselines on the evaluated datasets.
Our \textbf{contributions} are as follows:
\begin{itemize}
	\item We introduce a filtration-induced representation that keeps the original node set while replacing a single collapsed clique graph with multiple scale-separated co-occurrence views.
	\item We formulate unsupervised hypergraph alignment as a shared-coupling multi-scale GW problem, where one transport plan must preserve structural relationships across all filtration levels. The key distinction of \hypergwl is not a new GW solver, but the construction of filtration-induced relational channels from hyperedges and the use of one shared coupling to align them jointly.
	\item We provide controlled perturbation experiments and ablations showing when multi-scale filtration helps, and where sparse hypergraphs remain challenging. We show that \hypergwl{} is robust to structural noise and in almost all cases outperforms state-of-the-art baselines.
\end{itemize}

\section{Related Work}

\noindent
\textbf{Graph alignment.}
Graph alignment is closely related to graph isomorphism and the quadratic
assignment problem~\cite{lawler1963quadratic,yan2020learning}, and has been
studied across pattern recognition, computer vision, bioinformatics, and network
analysis~\cite{conte2004thirty,foggia2014graph,yan2016short,
	trung2020comparative,tang2025network}. 
We focus on the unrestricted setting, where correspondences are inferred from
topology alone~\cite{skitsas2023comprehensive}.
Representative methods include embedding-based approaches such as REGAL and
CONE~\cite{heimann2018regal,chen2020cone}, spectral methods such as
GRASP~\cite{hermanns2023grasp}, optimal-transport formulations such as
PARROT and SGWL~\cite{zeng2023parrot,xu2019scalable}, and relaxed QAP methods
such as FUGAL~\cite{bommakanti2024fugal}. 
These methods operate on ordinary graphs; applying them to hypergraphs requires
a graph representation, such as clique or bipartite expansion.

\smallskip
\noindent
\textbf{Hypergraph and higher-order alignment.}
Hypergraphs naturally model higher-order interactions, and hypergraph learning
has received increasing attention~\cite{gao2020hypergraph,ccatalyurek2023more,
	antelmi2023survey}. 
Hypergraph alignment, however, is less developed, especially without features,
labels, or seed matches. 
Prior work includes seed-based hypergraph manifold alignment~\cite{tan2014mapping},
tensor-based higher-order alignment such as TAME~\cite{mohammadi2016triangular},
and feature-driven hypergraph matching methods in computer vision
~\cite{nguyen2015flexible,liao2021hypergraph,zheng2024cursor}. 
These direct higher-order formulations can preserve hyperedge interactions more
explicitly, but they are often computationally demanding and less convenient for
non-uniform hypergraphs. 
Closest to our setting are HyperAlign~\cite{do2024unsupervised}, which learns
topology-derived hypergraph embeddings using contrastive and adversarial
training, and ELRUHNA~\cite{ibrahim2025elruhna}, which aligns vertices and
hyperedges through the bipartite incidence representation. 
In contrast, \hypergwl{} aligns the original node sets directly, uses no node or
hyperedge features, and replaces a single collapsed clique graph by
filtration-induced co-occurrence views.

\smallskip
\noindent
\textbf{Gromov--Wasserstein variants and multi-scale structure.}
Gromov--Wasserstein discrepancy aligns relational structures by preserving
pairwise distances or similarities~\cite{memoli2011gromov,peyre2016gromov}.
Fused and attributed variants incorporate node, edge, or multi-dimensional
relational features~\cite{vayer2020fused,yang2024fngw,kawano2024mfgw}, and
$Z$-Gromov--Wasserstein generalizes this view to vector-valued relational
kernels~\cite{bauer2025zgw}. 
Our formulation is complementary: the relational channels are not observed
attributes, but are generated from the hypergraph by a filtration. Thus,
\hypergwl{} can be interpreted as a filtration-induced vector-valued GW
objective whose channels encode how pairwise co-occurrence appears across
hyperedge scales.

\section{Preliminaries}
\label{sec:preliminaries}

We use $[k]$ with $k\in\mathbb{N}$ to denote the set
$\{1,\ldots,k\}$ and write
$
\Simplex{k}
=
\left\{
w\in\mathbb{R}_{\ge 0}^k
\mid
\sum_{m=1}^k w_m=1
\right\}
$
for the probability simplex.
An undirected hypergraph is a pair $G=(V,E)$, where $V$ is a finite set of
nodes and $E$ is a finite multiset over $2^V\setminus\{\emptyset\}$.
Each element $e\in E$ is a non-empty subset of nodes.
We write
$|e|$ for the cardinality of a hyperedge, i.e., the number of nodes incident
to it. A hypergraph is $k$-uniform if all hyperedges have cardinality $k$.
A $2$-uniform hypergraph is an ordinary graph.

\smallskip
\noindent
\textbf{Unsupervised hypergraph alignment.} Let $G_s=(V_s,E_s)$ and $G_t=(V_t,E_t)$ be two hypergraphs. For notational simplicity, assume first that $|V_s|=|V_t|$ and that there exists an unknown ground-truth bijection $ \tau:V_s\to V_t. $ The goal of unsupervised hypergraph alignment is to output an estimated bijection $ \phi:V_s\to V_t $ using only the observed hypergraph structures. In particular, we do not assume node features, hyperedge features, side information, labels, or known seed matches~\cite{do2024unsupervised,ibrahim2025elruhna}. If $|V_s|\neq |V_t|$, we pad the smaller node set with isolated dummy nodes so that a bijective alignment can be computed on equal-size node sets.

\smallskip
\noindent
\textbf{Graph representations of hypergraphs.}
A common way to apply graph-alignment methods to hypergraphs is to first
reduce the hypergraph to an ordinary graph. We consider two standard
reductions: the clique representation and bipartite representation~\cite{lee2024survey,zhou2006learning}.

The \emph{clique representation} keeps the original node set and connects two
nodes whenever they co-occur in at least one hyperedge. Formally, it maps
$G=(V,E)$ to the graph
$
K(G)=(V,E_K)$ where $E_K=\bigl\{\{u,v\}\subseteq V : u\neq v,\ \exists e\in E
\text{ with } u,v\in e\bigr\}.
$
This representation is compact and keeps the alignment problem on the original nodes, but it is lossy: it collapses each higher-order hyperedge into pairwise edges, so distinct hypergraphs can induce the same clique graph.

The \emph{bipartite representation} introduces one auxiliary node $w_e$ for
each hyperedge $e\in E$ and connects it to all incident nodes. Formally,
$B(G)=(V\cup W,F)$ with $W=\{w_e : e\in E\}$ and
$(u,w_e)\in F$ iff $u\in e$. This representation preserves the incidence
structure exactly, but increases the alignment problem from $|V|$ to
$|V|+|E|$ nodes.

\smallskip
\noindent
\textbf{Gromov--Wasserstein (GW) discrepancy.}
GW discrepancy compares two relational structures through
a transport plan between their nodes~\cite{peyre2016gromov,xu2019gromov,xu2019scalable}. Let
$
(C_s,\mu_s)\in
\mathbb{R}^{|V_s|\times |V_s|}
\times
\Simplex{|V_s|}
$
and
$
(C_t,\mu_t)\in
\mathbb{R}^{|V_t|\times |V_t|}
\times
\Simplex{|V_t|}
$
be measured dissimilarity matrices. The feasible transport polytope is
\[
\Pi(\mu_s,\mu_t)
=
\left\{
T\in\mathbb{R}_{\ge 0}^{|V_s|\times |V_t|}
\mid
T\mathbf{1}_{|V_t|}=\mu_s,\;
T^\top\mathbf{1}_{|V_s|}=\mu_t
\right\}.
\]
For a loss function $L$, the GW discrepancy is
\[
\min_{T\in\Pi(\mu_s,\mu_t)}
\sum_{i,k,j,\ell}
L\big(C_s[i,k],C_t[j,\ell]\big)
T_{ij}T_{k\ell}.
\]
In this work we use the squared loss
$
L(a,b)=(a-b)^2.
$

The transport plan $T$ is a soft relaxation of the desired node correspondence:
its entry $T_{ij}$ represents the mass assigned from source node $i$ to target
node $j$, with larger values indicating stronger support for matching the two
nodes. After optimizing $T$, we extract a discrete bijection by solving a linear
assignment problem, which selects the one-to-one correspondence with maximum
total support under the transport plan.

\begin{figure*}
	\centering
	\includegraphics[width=1\linewidth]{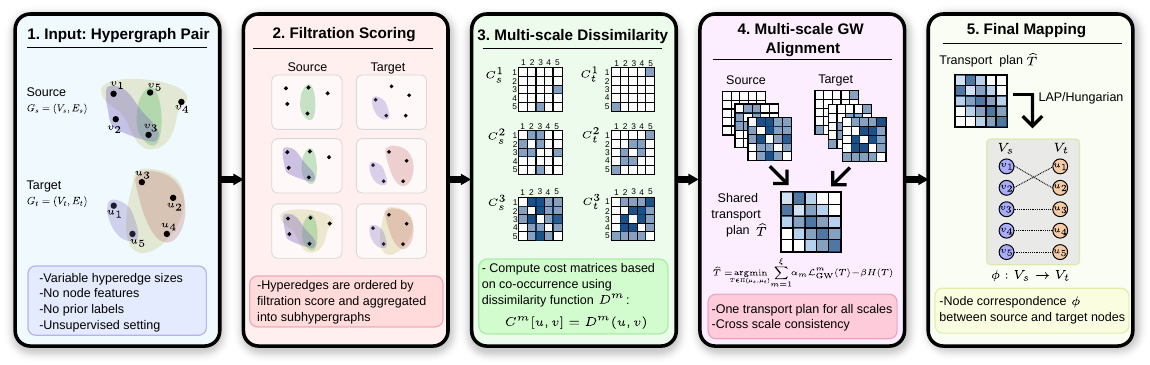}
	\caption{Overview of \hypergwl{}. Given two hypergraphs, \hypergwl assigns each hyperedge a filtration score and constructs subhypergraphs for the source and target hypergraphs. Each filtration subhypergraph is converted into a clique-induced co-occurrence dissimilarity matrix, producing multi-scale structural views for the source and target hypergraphs. A single shared Gromov--Wasserstein transport plan is then optimized across all scales, enforcing cross-scale consistency, and the final discrete node correspondence is obtained by solving a linear assignment problem on the optimized shared transport plan.}
	\label{fig:main2}
\end{figure*}

\section{Hypergraph Alignment Framework}
\label{sec:method}

We now derive our hypergraph alignment framework. The input consists of two
hypergraphs, while the desired output is a single node correspondence. Thus,
the method must decide both how to convert higher-order hypergraph structure
into comparable node-level quantities, and how to use these quantities to obtain
one globally consistent alignment.

\Cref{fig:main2} gives an overview of \hypergwl. Our starting point is
to keep the compactness of clique-based pairwise costs while avoiding a single
collapsed clique representation. \hypergwl instead reveals each hypergraph through a hyperedge filtration. At every filtration level, the active hyperedges are converted into a clique-induced
co-occurrence dissimilarity matrix. This yields a sequence of pairwise
structural views on the original node set. The filtration score determines the
order in which hyperedge evidence enters the representation; it may be based on
hyperedge size, node-degree structure, or other structural properties of the
hyperedge.

\hypergwl{} then aligns these views jointly, rather than independently, by optimizing
one shared Gromov--Wasserstein transport plan across all filtration levels. The
resulting coupling must explain the source and target structures consistently
across hyperedge scales, and is finally decoded into a discrete bijection by
solving a linear assignment problem. 

\subsection{From Hyperedges to Filtration-Induced Costs}
Let $G=(V,E)$ be a hypergraph. To obtain a multi-scale representation, we
assign each hyperedge a structural score through a function
$\omega:E\to\mathbb{R}$ and order the hyperedges by this score. The score can
encode different structural properties of a hyperedge. In this work, we use a
degree-aware score. Let $\deg(v)$ denote the number of hyperedges incident to
node $v$. We define
\begin{equation}\label{eq:deg}
	\omega_{\mathrm{deg}}(e) =
	\frac{\sum_{v \in e} \deg(v)}
	{\max_{e' \in E} \sum_{u \in e'} \deg(u)}.
\end{equation}

This score orders hyperedges by the normalized total incident degree of their
nodes. In increasing score order, it exposes
hyperedges from lower-degree, more peripheral regions toward hyperedges incident
to increasingly prominent nodes. Thus, it uses the node-degree structure to determine when each
hyperedge enters the multi-scale representation.

One could create a filtration level for every distinct hyperedge score. In
practice, however, this often produces many weakly informative levels:
consecutive levels may differ by only a few hyperedges, and some score changes
may affect only one of the two hypergraphs. Such one-sided or very small
refinements increase the cost of the multi-scale objective while contributing
little comparative evidence for a shared alignment.
We therefore construct a coarser set of synchronized filtration levels for the
source and target hypergraphs. The source and target score sequences are scanned
jointly in increasing order. A score range is retained only after both
hypergraphs have contributed at least one new hyperedge since the previous
retained range. 
This removes one-sided refinements and ensures
that every retained level activates new structure in both domains. The retained
ranges are then coarsened uniformly into a prescribed number of buckets. For
$x\in \{s,t\}$, let $B_x^1,\ldots,B_x^\xi$
denote the resulting buckets of hyperedges in hypergraph $G_x$, where $\xi$
controls the filtration resolution: smaller values yield coarser
representations, while larger values retain more synchronized score variation.

Each level $m$ is associated with a set of active hyperedges
$E_x^m\subseteq E_x$, determined by the chosen aggregation mode:
\[
E_x^m =
\begin{cases}
	\bigcup_{\ell=1}^{m} B_x^\ell, & \text{cumulative},\\
	B_x^m, & \text{non-cumulative}.
\end{cases}
\]
The cumulative mode accumulates
co-occurrence evidence across scales, whereas the non-cumulative mode isolates
the structural signal specific to each score range. Both modes induce a
subhypergraph $G_x^m=(V_x,E_x^m)$ at each level $m\in[\xi]$.

At each filtration level, we convert the active hyperedges into a node-pair
dissimilarity matrix. Let $D_x^m(u,v)$ denote the dissimilarity between nodes
$u$ and $v$ induced by the active hyperedges in $E_x^m$. We set
\begin{equation}
	C_x^m[u,v] = D_x^m(u,v), \qquad u\neq v,
\end{equation}
and $C_x^m[u,u]=0$. Concretely, we use the binary-overlap dissimilarity. Let
$\delta_x^m(u,v)$ be the number of active hyperedges in $E_x^m$ that contain
both $u$ and $v$. Then
\begin{equation}\label{eq:overlap}
	D_{x,\mathrm{bin}}^m(u,v)
	=
	\begin{cases}
		0, & \delta_x^m(u,v) > 0,\\
		1, & \delta_x^m(u,v) = 0.
	\end{cases}
\end{equation}
Thus, two nodes are close at level $m$ if they co-occur in at least one active
hyperedge. This binary-overlap construction follows the robustness intuition
used in hypergraph structure learning, where noisy incidences are often handled
through thresholding, sparsification, or binary masks~\cite{cai2022hypergraph,
	zhang2022deep}. Here, thresholding the co-occurrence $\delta_x^m(u,v)$
keeps only binary pairwise evidence, making the cost insensitive to
multiplicity perturbations that do not change whether a pair co-occurs.

Applying this construction to both hypergraphs yields the 
dissimilarity matrices
$
\{C_s^m\}_{m=1}^{\xi}
$
and
$
\{C_t^m\}_{m=1}^{\xi}.
$

\subsection{Multi-Scale Optimal Transport}
\label{sec:shared-coupling-motivation}

The filtration gives several structural views of each hypergraph. The alignment
problem, however, asks for one correspondence between the source nodes and the
target nodes. Therefore, the transport plan should not depend on the filtration
level: if source node $i$ corresponds to target node $j$, that correspondence
should be globally meaningful across scales.

This leads to a shared-coupling formulation. We seek a single transport plan
$
T\in\Pi(\mu_s,\mu_t).
$
The marginals $\mu_s$ and $\mu_t$ are normalized node-degree distributions,
where the degree of a node is its number of incident hyperedges.

At one filtration level $m$, the Gromov--Wasserstein distortion of a coupling
$T$ is
\begin{equation}
	\label{eq:single-scale-gw}
	\mathcal{L}_{\mathrm{GW}}^m(T)
	=
	\sum_{i,k,j,\ell}
	\left(
	C_s^m[i,k]-C_t^m[j,\ell]
	\right)^2
	T_{ij}T_{k\ell}.
\end{equation}
This measures whether pairs of source nodes that are close at level $m$ are
transported to pairs of target nodes that are also close at level $m$.

Since no single filtration level is privileged, the natural multi-scale objective
is to aggregate these distortions:
\begin{equation}
	\label{eq:joint-msgw}
	\widehat T
	=
	\arg\min_{T\in\Pi(\mu_s,\mu_t)}
	\sum_{m=1}^{\xi}
	\alpha_m
	\mathcal{L}_{\mathrm{GW}}^m(T)
	-
	\beta H(T),
\end{equation}
where
$
\alpha_m\ge 0,$ with $\sum_{m=1}^{\xi}\alpha_m=1,
$
and
\[
H(T)=-\sum_{i,j}T_{ij}(\log T_{ij}-1)
\]
is the entropic regularizer.

Equation~\eqref{eq:joint-msgw} is the central objective of \hypergwl{}.

\begin{algorithm2e}[t]\small
	\caption{\textnormal{\hypergwl}}
	\label[algorithm]{alg:hypergwl}
	\KwIn{
		Hypergraphs $G_s=(V_s,E_s)$ and $G_t=(V_t,E_t)$;
		filtration score $\omega$; co-occurrence dissimilarity $D$;
		aggregation mode $\in\{\textnormal{cumulative},\textnormal{non-cumulative}\}$;
		entropic weight $\beta>0$;
		number of buckets $\xi$
	}
	\KwOut{Bijective node mapping $\phi:V_s\to V_t$}
	
	Assign all hyperedges in $E_s$ and $E_t$ a filtration score using $\omega$\;
	
	Build synchronized bucket pairs
	$
	\{(B_s^m,B_t^m)\}_{m=1}^{\xi}
	$
	from the source and target hyperedge scores\;
	
	Build node marginals $\mu_s,\mu_t$\;
	
	Construct active hyperedge sets
	$
	\{(E_s^m,E_t^m)\}_{m=1}^{\xi}
	$
	from the bucket pairs using the chosen aggregation mode\;
	
	Construct %
	dissimilarity matrices
	$
	\{(C_s^m,C_t^m)\}_{m=1}^{\xi}
	$
	from the active hyperedge sets using $D$\;
	
	Compute scale weights $\alpha\in\Simplex{\xi}$ from
	$
	\{(B_s^m,B_t^m)\}_{m=1}^{\xi}
	$
	
	Initialize $T\leftarrow \mu_s\mu_t^\top$\;
	
	Solve the joint multi-scale GW (\Cref{eq:joint-msgw}) to obtain
	$\widehat T$\;
	
	Solve a LAP on $-\widehat T$ to obtain
	$\phi:V_s\to V_t$\;
	
	\Return{$\phi$}
\end{algorithm2e}

\subsection{The \hypergwl{} Algorithm}
\label{sec:algorithm}

\Cref{alg:hypergwl} summarizes \hypergwl{}.
We optimize Equation~\eqref{eq:joint-msgw} using the standard entropic GW
scheme: each outer iteration linearizes the quadratic GW distortion around the
current coupling and solves the resulting entropic optimal-transport subproblem
by Sinkhorn scaling; see, e.g.,
\cite{peyre2016gromov,peyre2019computational}. Starting from
$
T^{(0)}=\mu_s\mu_t^\top,
$
each filtration level $m$ yields a linearized cost $M_m(T^{(q)})$ at the current
coupling. The multi-scale update combines these costs before the transport
update,
\[
M(T^{(q)})
=
\sum_{m=1}^{\xi}\alpha_m M_m(T^{(q)}).
\]

After convergence, \hypergwl{} decodes a discrete node correspondence by solving
a linear assignment problem on the optimized transport plan using the Hungarian
algorithm~\cite{kuhn1955hungarian}. 

\smallskip
\noindent
\textbf{Weighting rule:}
We choose the weights $\alpha_m$ for $m\in[\xi]$ data-dependently via \emph{balanced edge-count weighting}. First we set 
\begin{equation}\label{eq:weight}
	w_m =
	\sqrt{|B_m^s|\,|B_m^t|},
\end{equation}
where $B_m^s$ and $B_m^t$ are the source and target hyperedges introduced by
the $m$-th synchronized bucket. We then use
$\alpha_m = w_m / \sum_{\ell=1}^{\xi} w_\ell$.
Thus, a level receives high weight only when
both hypergraphs contribute substantial new hyperedge evidence.

\subsection{Interpretation}
\label{sec:joint-interpretation}

The shared coupling in Equation~\eqref{eq:joint-msgw} enforces cross-scale
consistency. A candidate correspondence receives high support only if it helps
preserve source-target structural relationships across the filtration. In other
words, \hypergwl{} does not ask which filtration level is best. Instead, it asks
which single node correspondence best explains the entire sequence of structural
views.
The weights $\alpha_m$ determine how strongly each filtration level contributes.
Uniform weights treat all selected scales equally, while data-dependent weights
can emphasize levels that introduce more balanced or more substantial structural
evidence in the two hypergraphs.

\smallskip
\noindent
\textbf{Filtration-induced vector-valued GW view.} The multi-scale objective also admits a useful interpretation through the lens
of vector-valued GW. The derivation above treats
\(\{C_s^m,C_t^m\}_{m=1}^{\xi}\) as filtration-induced structural views and
jointly optimizes one coupling across all of them. Equivalently, we may regard
the filtration levels as relational channels attached to each node pair.
For every source node pair \((i,k)\), define the multi-scale relational
signature
\[
r_s(i,k)
=
\big(
C_s^1[i,k],\ldots,C_s^\xi[i,k]
\big)
\in \mathbb{R}^{\xi},
\]
and define \(r_t(j,\ell)\) analogously for the target. Equip
\(\mathbb{R}^{\xi}\) with the weighted squared distance
\[
d_{\alpha}(a,b)^2
=
\sum_{m=1}^{\xi}\alpha_m(a_m-b_m)^2 .
\]
Then the distortion term in Equation~\eqref{eq:joint-msgw} can be written as
\[
\sum_{i,k,j,\ell}
d_{\alpha}\big(r_s(i,k),r_t(j,\ell)\big)^2
T_{ij}T_{k\ell}.
\]
Equivalently, by absorbing the weights into the coordinates,
\[
c_s(i,k)
=
\big(
\sqrt{\alpha_1}C_s^1[i,k],
\ldots,
\sqrt{\alpha_\xi}C_s^\xi[i,k]
\big),
\]
we obtain
$
d_{\alpha}\big(r_s(i,k),r_t(j,\ell)\big)^2
=
\|c_s(i,k)-c_t(j,\ell)\|_2^2 .
$

Thus, \hypergwl can be viewed as a filtration-induced vector-valued GW
objective: it combines several structural channels, one per filtration level,
into a single relational kernel that is preserved by one shared coupling.
This perspective connects our
objective to fused and vector-valued GW formulations, while highlighting the
specific role of the hypergraph filtration. In attributed GW, the channels are
typically observed node or edge features; here they are generated from
higher-order hyperedges and record how pairwise co-occurrence structure appears
across scales. The connection to 
vector-valued (and fused) GW clarifies the form of the objective, while the hypergraph
contribution lies in how the channels are constructed: they are not observed
node features or edge attributes, but co-occurrence views induced by a hyperedge
filtration, aggregated either cumulatively or per-scale depending on the chosen
mode~\cite{vayer2020fused,kawano2024mfgw}.

\smallskip
\noindent
\textbf{Consistency in the noiseless case.}
Suppose $G_s$ and $G_t$ are related by a node permutation $\tau$, and that the
filtration score, bucket construction, aggregation mode, and dissimilarity
function are all relabeling-invariant. Then the source and target score
multisets coincide, so the synchronized bucket construction selects
corresponding levels in the two hypergraphs. Consequently, $\tau$ maps each
source active-edge set to its target counterpart, and the induced cost matrices
satisfy $C_t^m = P_\tau^\top C_s^m P_\tau$ for all $m \in [\xi]$. The
ground-truth permutation therefore attains zero multi-scale GW distortion when
$\beta=0$, and since the objective is a nonnegative weighted sum of squares, it
is a global minimizer. However, this does not make \hypergwl a hypergraph isomorphism test:
the filtration-induced co-occurrence matrices are pairwise summaries of
higher-order structure and are in general lossy, so distinct hypergraphs can
induce identical views. It shows only that the objective is consistent with
exact structural alignment. For \(\beta>0\), the entropic solution is a smoothed relaxation; as
\(\beta\to0\), its limit points lie among the zero-distortion minimizers.
If the zero-distortion coupling is unique, the solution concentrates on it.\looseness=-1

\smallskip
\noindent
\textbf{Practical interpretation.}
In noisy settings, \hypergwl searches for one coupling that jointly explains all
filtration-induced views. A correspondence is supported only when it preserves
node-pair relationships across multiple structural scales, making the objective
more restrictive than aligning one collapsed clique representation while still
avoiding the auxiliary hyperedge nodes of bipartite expansion. Cumulative
aggregation reinforces co-occurrence patterns as evidence accumulates across the
filtration, whereas non-cumulative aggregation isolates the signal specific to
each score range and avoids carrying noisy incidences into later levels.

\subsection{Complexity}
\label{sec:complexity}

Let $|V_s|=|V_t|=n$, $|E|=\max(|E_s|,|E_t|)$, $\xi$ be the number of
filtration buckets (usually a small constant), and $K$ be the number of outer iterations of
the entropic GW solver.

Selecting synchronized buckets requires sorting the hyperedge scores and costs
$O(|E|\log |E|)$ time. Constructing the filtration-induced cost matrices requires
expanding each hyperedge $e$ over all node pairs it contains, which costs
$O(|e|^2)$. Thus, over both hypergraphs, the total preprocessing cost is
$
O(
|E|\log |E|
+
\sum_{x\in\{s,t\}}\sum_{e\in E_x}|e|^2
)
$.

The joint multi-scale GW solver evaluates all $\xi$ cost-matrix pairs at each
outer iteration. With dense matrix operations, this costs $O(\xi K n^3)$. The
final linear assignment step costs $O(n^3)$, which is dominated by the GW term.
Hence the total dense running time is
$
O\left(
|E|\log |E|
+
\sum_{x\in\{s,t\}}\sum_{e\in E_x}|e|^2
+
\xi K n^3
\right).
$
The space complexity is $O(\xi n^2)$ for the scale-specific cost matrices plus
the dense quantities used by the shared GW solver.

\section{Experiments}\label{sec:experiments}
We study the following research questions:
\begin{itemize}
	\item \textbf{RQ1:}
	How does \hypergwl compare with graph- and hypergraph-alignment baselines
	under increasing structural perturbation?
	
	\item \textbf{RQ2:}
	Does \hypergwl benefit from jointly using multiple filtration-induced views, and
	are its default degree-aware filtration and balanced scale weights important? What is the impact of the number of buckets $\xi$?
	
	\item \textbf{RQ3:}
	What is the computational cost of \hypergwl compared to the baselines?
\end{itemize}

\begin{table}[t]\centering
	\caption{Dataset statistics.}\label{table:stats}
	\renewcommand{\arraystretch}{1.0}
	\setlength{\tabcolsep}{10pt}
	\resizebox{1\linewidth}{!}{
		\begin{tabular}{lrrrrr}
			\toprule
			\textbf{Dataset} & $|V|$ & $|E|$ &  \textbf{Max~$|e|$} & \textbf{Avg.~$|e|$} & \textbf{Avg.~deg.} \\
			\midrule                 
			\emph{NDC}        & 628 &  796  &   39  &  7.20   &   9.12   \\
			\emph{Email}        & 986 &  24\,520  &   40  &  3.62   &  90.04  \\
			\emph{House}	& 1\,494 & 54\,933  & 399 & 22.14 & 814.39 \\
			\emph{Dawn}        & 2\,290 &  138\,742   &   16  &   3.99  &  241.55  \\
			\bottomrule
	\end{tabular}}
\end{table}

\smallskip
\noindent
\underline{\textbf{Datasets:}}
We use the following real-world hypergraphs: 
\begin{itemize}
	\item \emph{NDC:} Drugs are hyperedges; nodes are their assigned class labels, from the National Drug Code Directory~\cite{Benson-2018-simplicial}.
	\item \emph{Email:} Nodes are email addresses; hyperedges are emails sent within a European research institution~\cite{Yin-2017-local}.
	\item \emph{House:} Nodes are members of the U.S. House; hyperedges are bill sponsor–cosponsor groups~\cite{Fowler-2006-cosponsorship}.
	\item \emph{Dawn:} Nodes are drugs; hyperedges are sets of drugs taken before emergency room visits~\cite{Amburg-2020-categorical}.
\end{itemize}
\Cref{table:stats} shows the dataset statistics.
Following standard alignment evaluation practice~\cite{xu2019scalable,bommakanti2024fugal,ibrahim2025elruhna}, we construct semi-synthetic benchmark families from the real-world hypergraphs using controlled structural perturbations and random node relabeling. Specifically, we use the following two benchmarks:

\emph{(i) Incidence noise benchmark:}
Starting from a single observed hypergraph $G_s$, we generate a structurally similar but distinct hypergraph $G_t$ by applying random incidence noise followed by a uniformly random node relabeling. The noise model operates on the incidence matrix: each incidence-matrix entry is independently flipped with probability $p$, adding or removing node memberships from hyperedges. This models scenarios where two hypergraphs represent the same latent system observed through different (imperfect) channels, e.g., co-authorship networks extracted from different
bibliographic databases, or email group communications recorded by different logging systems. The parameter $p$ controls the structural dissimilarity between $G_s$ and $G_t$.

\emph{(ii) Subgraph-sampling benchmark:} Additionally, we evaluate a partial-observation setting. Starting from a latent hypergraph $G$, we independently sample the source and target hypergraphs by retaining each hyperedge with probability $1-p$, and then apply a random node permutation to the target. Unlike incidence flips, this perturbation does not alter the internal composition of retained hyperedges; instead, it models two incomplete snapshots of the same underlying higher-order system. This setting is common when hyperedges correspond to observed group events, such as emails, bills, prescriptions, or co-occurrences, where different data sources may observe different subsets of interactions.

For both benchmark families and each dataset, we generate hypergraph pairs $(G_s, G_t)$ at noise levels $p \in \{0.05, 0.10, 0.15, 0.20, 0.25\}$, with 10 independent random trials per level. Each instance comes with a ground-truth node correspondence $\tau$ (the permutation used to relabel the target nodes), so alignment quality is measured as the fraction of correctly recovered node correspondences.

\begin{table*}[htbp]
	\centering
	\caption{Accuracy (\%) by dataset and noise level for the two benchmarks. Each entry reports mean $\pm$ standard deviation over ten independent runs. OOT denotes out of time and OOM denotes out of memory. }
	\label{tab:accuracy}
	\begin{subtable}[t]{1\linewidth}
		\renewcommand{\arraystretch}{1.0}
		\setlength{\tabcolsep}{3pt}
		\resizebox{\textwidth}{!}{%
			\begin{tabular}{llcccccccccccc}
				\toprule
				\textbf{Dataset} & \textbf{Noise} & {SGWL cli.} & {SGWL bip.} & {PARROT cli.} & {PARROT bip.} & {FUGAL cli.} & {FUGAL bip.} & {BIGALIGN} & {TAME} & {HCN+CONE} & {ELRUHNA} & {\hypergwlc} & {\hypergwlnc} \\
				\midrule
				\emph{NDC} 
				& $p=0.00$ & 59.3 $\pm$ 2.9 & 45.7 $\pm$ 1.4 & 53.8 $\pm$ 1.7 & 34.1 $\pm$ 0.0 & 61.2 $\pm$ 2.2 & 29.0 $\pm$ 1.1 & 26.6 $\pm$ 0.0 & 35.4 $\pm$ 3.2 & 46.3 $\pm$ 2.4 & \textbf{66.6} $\pm$ 2.6  & {66.2} $\pm$ 3.2 & 64.8 $\pm$ 1.0\\
				& $p=0.05$ & 33.2 $\pm$ 2.5 & 26.8 $\pm$ 3.1 & 1.7 $\pm$ 0.6 & 14.3 $\pm$ 1.1 & 44.6 $\pm$ 1.7 & 26.9 $\pm$ 1.4 & 11.5 $\pm$ 0.9 & 17.9 $\pm$ 5.9 & 26.2 $\pm$ 2.1 & \textbf{58.4} $\pm$ 2.5 & 55.3 $\pm$ 4.3 & 41.9 $\pm$ 3.7 \\
				& $p=0.10$ & 22.0 $\pm$ 1.1 & 16.7 $\pm$ 1.8 & 0.5 $\pm$ 0.2 & 7.8 $\pm$ 0.7 & 35.7 $\pm$ 3.1 & 24.3 $\pm$ 1.2 & 6.0 $\pm$ 1.4 & 12.1 $\pm$ 5.6 & 19.2 $\pm$ 2.2 & \textbf{49.9} $\pm$ 2.6 & 47.1 $\pm$ 2.0 & 28.7 $\pm$ 2.5 \\
				& $p=0.15$ & 15.7 $\pm$ 2.3 & 11.2 $\pm$ 1.3 & 0.2 $\pm$ 0.1 & 5.3 $\pm$ 0.7 & 13.9 $\pm$ 5.0 & 22.3 $\pm$ 1.3 & 4.0 $\pm$ 0.7 & 8.6 $\pm$ 3.7 & 12.0 $\pm$ 2.8 & 32.4 $\pm$ 10.9 & \textbf{40.9} $\pm$ 1.6 & 20.9 $\pm$ 1.4 \\
				& $p=0.20$ & 10.1 $\pm$ 3.1 & 7.0 $\pm$ 0.9 & 0.2 $\pm$ 0.1 & 3.3 $\pm$ 0.7 & 3.5 $\pm$ 2.5 & 20.7 $\pm$ 1.6 & 3.1 $\pm$ 0.4 & 7.9 $\pm$ 4.1 & 9.1 $\pm$ 2.1 & 25.7 $\pm$ 9.2 & \textbf{35.5} $\pm$ 1.4 & 16.1 $\pm$ 2.1 \\
				& $p=0.25$ & 6.4 $\pm$ 1.2 & 5.1 $\pm$ 0.6 & 0.2 $\pm$ 0.1 & 2.6 $\pm$ 0.5 & 0.2 $\pm$ 0.1 & 18.5 $\pm$ 1.8 & 2.2 $\pm$ 0.5 & 4.4 $\pm$ 3.7 & 6.8 $\pm$ 1.9 & 18.1 $\pm$ 6.7 & \textbf{30.6} $\pm$ 1.1 & 11.6 $\pm$ 1.1 \\
				\midrule
				\emph{Email} 
				& $p=0.00$ & 98.4 $\pm$ 0.2 & OOT & 95.1 $\pm$ 0.3 & 86.1 $\pm$ 0.0 & 40.9 $\pm$ 0.8 & OOT & 25.1 $\pm$ 0.3 & OOT & 83.5 $\pm$ 1.8 &  \textbf{99.8} $\pm$ 0.1  & {98.9} $\pm$ 0.2 & 98.7 $\pm$ 0.3\\
				& $p=0.05$ & 67.9 $\pm$ 1.3 & OOT & 0.2 $\pm$ 0.1 & 6.6 $\pm$ 0.5 & 76.1 $\pm$ 5.0 & OOT & 0.8 $\pm$ 0.2 & OOT & 37.4 $\pm$ 5.6 & 78.0 $\pm$ 18.7 & {94.2} $\pm$ 0.8 & \textbf{94.6} $\pm$ 0.4 \\
				& $p=0.10$ & 56.6 $\pm$ 2.4 & OOT & 0.2 $\pm$ 0.1 & 1.1 $\pm$ 0.1 & 0.3 $\pm$ 0.4 & OOT & 0.3 $\pm$ 0.1 & OOT & 11.8 $\pm$ 3.6 & 63.1 $\pm$ 27.9 & {89.0} $\pm$ 0.8 & \textbf{92.7} $\pm$ 0.7 \\
				& $p=0.15$ & 49.2 $\pm$ 1.1 & OOT & 0.1 $\pm$ 0.0 & 0.6 $\pm$ 0.1 & 0.0 $\pm$ 0.0 & OOT & 0.2 $\pm$ 0.1 & OOT & 3.5 $\pm$ 1.3 & 58.6 $\pm$ 33.6 & {86.5} $\pm$ 0.7 & \textbf{89.0} $\pm$ 0.7 \\
				& $p=0.20$ & 44.1 $\pm$ 1.3 & OOT & 0.2 $\pm$ 0.1 & 0.4 $\pm$ 0.1 & 0.0 $\pm$ 0.1 & OOT & 0.2 $\pm$ 0.1 & OOT & 2.1 $\pm$ 0.9 & 60.2 $\pm$ 28.9 & {84.7} $\pm$ 0.6 & \textbf{86.0} $\pm$ 1.0 \\
				& $p=0.25$ & 0.2 $\pm$ 0.1 & OOT & 0.0 $\pm$ 0.1 & 0.3 $\pm$ 0.1 & 0.0 $\pm$ 0.0 & OOT & 0.2 $\pm$ 0.1 & OOT & 1.6 $\pm$ 1.0 & 37.4 $\pm$ 34.7 & \textbf{82.5} $\pm$ 0.7 & 82.4 $\pm$ 1.0 \\
				\midrule
				\emph{House} 
				& $p=0.00$ & 0.1 $\pm$ 0.1 & OOT & 64.7 $\pm$ 0.1 & 99.9 $\pm$ 0.0 & 34.6 $\pm$ 0.5 & OOT & OOT & OOT & 0.3 $\pm$ 0.3 & OOT & \textbf{100.0} $\pm$ 0.0 & \textbf{100.0} $\pm$ 0.0\\
				& $p=0.05$ & 0.1 $\pm$ 0.0 & OOT & 0.2 $\pm$ 0.0 & 1.5 $\pm$ 0.0 & 0.0 $\pm$ 0.0 & OOT & OOT & OOT & 0.2 $\pm$ 0.2 & OOT & {98.5} $\pm$ 0.2 & \textbf{99.0} $\pm$ 0.0 \\
				& $p=0.10$ & 0.1 $\pm$ 0.0 & OOT & 0.1 $\pm$ 0.0 & 1.0 $\pm$ 0.0 & 0.0 $\pm$ 0.0 & OOT & OOT & OOT & 0.1 $\pm$ 0.1 & OOT & {94.5} $\pm$ 0.2 & \textbf{96.2} $\pm$ 0.3 \\
				& $p=0.15$ & 0.1 $\pm$ 0.0 & OOT & 0.1 $\pm$ 0.0 & 0.5 $\pm$ 0.0 & 0.0 $\pm$ 0.0 & OOT & OOT & OOT & 0.1 $\pm$ 0.1 & OOT & {87.1} $\pm$ 0.4 & \textbf{89.0} $\pm$ 1.1 \\
				& $p=0.20$ & 0.0 $\pm$ 0.0 & OOT & 0.0 $\pm$ 0.0 & 0.5 $\pm$ 0.0 & 0.0 $\pm$ 0.0 & OOT & OOT & OOT & 0.1 $\pm$ 0.1 & OOT & {76.9} $\pm$ 0.5 & \textbf{79.2} $\pm$ 1.7 \\
				& $p=0.25$ & 0.0 $\pm$ 0.0 & OOT & 0.0 $\pm$ 0.0 & 0.4 $\pm$ 0.0 & 0.0 $\pm$ 0.0 & OOT & OOT & OOT & 0.1 $\pm$ 0.1 & OOT & {61.4} $\pm$ 1.6 & \textbf{67.9} $\pm$ 1.2 \\
				\midrule
				\emph{Dawn} 
				& $p=0.00$ & 0.1 $\pm$ 0.0 & OOM & 93.1 $\pm$ 0.3 & OOM & 51.9 $\pm$ 1.6 & OOM & OOM & OOT & 0.2 $\pm$ 0.3 &  OOT &  \textbf{98.3} $\pm$ 0.0 & \textbf{98.3} $\pm$ 0.0 \\
				& $p=0.05$ & 0.0 $\pm$ 0.0 & OOM & 0.0 $\pm$ 0.0 & OOM & 0.1 $\pm$ 0.1 & OOM & OOM & OOT & 0.1 $\pm$ 0.1 & OOT & {62.5} $\pm$ 0.7 & \textbf{72.5} $\pm$ 0.0 \\
				& $p=0.10$ & 0.0 $\pm$ 0.1 & OOM & 0.0 $\pm$ 0.0 & OOM & 0.1 $\pm$ 0.0 & OOM & OOM & OOT & 0.0 $\pm$ 0.0 & OOT & {48.4} $\pm$ 0.3 & \textbf{61.9} $\pm$ 0.7 \\
				& $p=0.15$ & 0.0 $\pm$ 0.0 & OOM & 0.0 $\pm$ 0.0 & OOM & 0.0 $\pm$ 0.0 & OOM & OOM & OOT & 0.1 $\pm$ 0.1 & OOT & {38.4} $\pm$ 0.6 & \textbf{54.2} $\pm$ 0.6 \\
				& $p=0.20$ & 0.0 $\pm$ 0.0 & OOM & 0.0 $\pm$ 0.0 & OOM & 0.0 $\pm$ 0.0 & OOM & OOM & OOT & 0.1 $\pm$ 0.0 & OOT & {30.5} $\pm$ 0.4 & \textbf{47.7} $\pm$ 0.7 \\
				& $p=0.25$ & 0.1 $\pm$ 0.1 & OOM & 0.0 $\pm$ 0.0 & OOM & 0.0 $\pm$ 0.0 & OOM & OOM & OOT & 0.0 $\pm$ 0.0 & OOT & {22.4} $\pm$ 0.8 & \textbf{40.2} $\pm$ 0.8 \\
				\bottomrule
			\end{tabular}
		}
		\caption{Results for the incidence noise benchmark.}
		\label{tab:accuracy-summary}
	\end{subtable}
	
	\begin{subtable}[t]{1\linewidth}
		\centering
		\renewcommand{\arraystretch}{1.0}
		\setlength{\tabcolsep}{3pt}
		\resizebox{\textwidth}{!}{%
			\begin{tabular}{llcccccccccccc}
				\toprule
				\textbf{Dataset} & \textbf{Noise} & {SGWL cli.} & {SGWL bip.} & {PARROT cli.} & {PARROT bip.} & {FUGAL cli.} & {FUGAL bip.} & {BIGALIGN} & {TAME} & {HCN+CONE} & {ELRUHNA} & {\hypergwlc} & {\hypergwlnc} \\
				\midrule
				\emph{NDC} & $p=0.05$ & 52.5 $\pm$ 3.1 & 40.3 $\pm$ 3.0 & 35.1 $\pm$ 3.9 & 19.8 $\pm$ 1.6 & 47.6 $\pm$ 19.0 & 22.8 $\pm$ 9.2 & 14.9 $\pm$ 1.6 & 33.6 $\pm$ 9.1 & 38.3 $\pm$ 4.0 & 55.4 $\pm$ 5.6 & \textbf{60.5} $\pm$ 1.7 & 48.1 $\pm$ 2.8 \\
				& $p=0.10$ & 45.5 $\pm$ 2.3 & 36.5 $\pm$ 2.3 & 22.9 $\pm$ 3.0 & 14.1 $\pm$ 1.3 & 44.3 $\pm$ 9.9 & 22.7 $\pm$ 5.4 & 11.0 $\pm$ 1.4 & 24.0 $\pm$ 7.3 & 32.4 $\pm$ 4.1 & 46.6 $\pm$ 7.5 & \textbf{53.2} $\pm$ 2.8 & 33.9 $\pm$ 6.8 \\
				& $p=0.15$ & 40.1 $\pm$ 2.8 & 31.7 $\pm$ 1.8 & 17.7 $\pm$ 2.8 & 11.1 $\pm$ 1.5 & 38.6 $\pm$ 14.4 & 19.8 $\pm$ 7.5 & 8.5 $\pm$ 1.2 & 20.1 $\pm$ 4.6 & 27.3 $\pm$ 4.3 & 43.1 $\pm$ 6.5 & \textbf{47.0} $\pm$ 3.0 & 28.1 $\pm$ 7.5 \\
				& $p=0.20$ & 37.2 $\pm$ 3.0 & 28.2 $\pm$ 2.1 & 15.6 $\pm$ 2.8 & 8.1 $\pm$ 0.9 & 23.9 $\pm$ 14.3 & 12.8 $\pm$ 7.9 & 6.4 $\pm$ 1.9 & 18.9 $\pm$ 5.7 & 22.2 $\pm$ 4.3 & 36.5 $\pm$ 6.6 & \textbf{43.9} $\pm$ 3.3 & 26.2 $\pm$ 5.1 \\
				& $p=0.25$ & 30.4 $\pm$ 2.6 & 23.6 $\pm$ 2.4 & 10.0 $\pm$ 1.1 & 6.6 $\pm$ 1.0 & 11.8 $\pm$ 11.9 & 6.1 $\pm$ 5.8 & 4.6 $\pm$ 1.0 & 13.2 $\pm$ 4.2 & 17.2 $\pm$ 3.9 & 32.7 $\pm$ 3.6 & \textbf{38.4} $\pm$ 3.8 & 15.8 $\pm$ 4.3 \\
				\midrule
				\emph{Email} & $p=0.05$ & 97.1 $\pm$ 0.7 & OOT & 49.4 $\pm$ 7.8 & 31.1 $\pm$ 1.3 & 38.6 $\pm$ 2.6 & OOT & 8.6 $\pm$ 0.8 & OOT & 78.3 $\pm$ 3.4 & 97.3 $\pm$ 0.4 & \textbf{98.1} $\pm$ 0.6 & 97.7 $\pm$ 0.7 \\
				& $p=0.10$ & 95.2 $\pm$ 1.1 & OOT & 32.6 $\pm$ 3.9 & 21.5 $\pm$ 1.6 & 35.3 $\pm$ 4.5 & OOT & 6.8 $\pm$ 0.8 & OOT & 71.8 $\pm$ 5.4 & 76.2 $\pm$ 27.7 & \textbf{97.0} $\pm$ 0.7 & 95.4 $\pm$ 1.8 \\
				& $p=0.15$ & 92.5 $\pm$ 1.6 & OOT & 22.4 $\pm$ 1.0 & 16.7 $\pm$ 1.3 & 36.1 $\pm$ 4.1 & OOT & 4.9 $\pm$ 0.6 & OOT & 61.3 $\pm$ 3.1 & 92.4 $\pm$ 1.2 & \textbf{95.5} $\pm$ 0.7 & 94.4 $\pm$ 0.7 \\
				& $p=0.20$ & 90.6 $\pm$ 1.4 & OOT & 16.6 $\pm$ 3.5 & 14.2 $\pm$ 1.2 & 51.9 $\pm$ 25.9 & OOT & 4.0 $\pm$ 0.6 & OOT & 55.0 $\pm$ 5.4 & 93.1 $\pm$ 0.7 & \textbf{94.5} $\pm$ 0.8 & 92.2 $\pm$ 0.8 \\
				& $p=0.25$ & 89.7 $\pm$ 1.8 & OOT & 14.6 $\pm$ 2.0 & 10.9 $\pm$ 1.0 & 76.4 $\pm$ 16.8 & OOT & 3.9 $\pm$ 0.7 & OOT & 52.8 $\pm$ 6.6 & 86.3 $\pm$ 5.4 & \textbf{93.5} $\pm$ 0.7 & 89.2 $\pm$ 2.4 \\
				\midrule
				\emph{House} & $p=0.05$ & 0.1 $\pm$ 0.1 & OOT & 45.7 $\pm$ 2.4 & 48.8 $\pm$ 1.1 & 20.5 $\pm$ 0.8 & OOT & OOT & OOT & 1.3 $\pm$ 1.6 & OOT & \textbf{100.0} $\pm$ 0.0 & \textbf{100.0} $\pm$ 0.0 \\
				& $p=0.10$ & 0.1 $\pm$ 0.1 & OOT & 38.6 $\pm$ 2.6 & 35.9 $\pm$ 2.5 & 15.9 $\pm$ 1.4 & OOT & OOT & OOT & 0.4 $\pm$ 0.4 & OOT & \textbf{100.0} $\pm$ 0.0 & \textbf{100.0} $\pm$ 0.0 \\
				& $p=0.15$ & 0.1 $\pm$ 0.1 & OOT & 35.6 $\pm$ 1.6 & 28.4 $\pm$ 0.3 & 15.1 $\pm$ 0.7 & OOT & OOT & OOT & 0.6 $\pm$ 0.7 & OOT & \textbf{100.0} $\pm$ 0.0 & \textbf{100.0} $\pm$ 0.0 \\
				& $p=0.20$ & 0.0 $\pm$ 0.0 & OOT & 31.2 $\pm$ 2.3 & 24.1 $\pm$ 2.2 & 13.3 $\pm$ 1.1 & OOT & OOT & OOT & 0.5 $\pm$ 0.6 & OOT & \textbf{100.0} $\pm$ 0.0 & 99.9 $\pm$ 0.1 \\
				& $p=0.25$ & 0.1 $\pm$ 0.1 & OOT & 27.7 $\pm$ 2.0 & 21.1 $\pm$ 1.1 & 10.7 $\pm$ 1.5 & OOT & OOT & OOT & 0.2 $\pm$ 0.0 & OOT & \textbf{100.0} $\pm$ 0.0 & 99.8 $\pm$ 0.2 \\
				\midrule
				\emph{Dawn} & $p=0.05$ & 0.0 $\pm$ 0.1 & OOM & 35.1 $\pm$ 1.3 & OOM & 18.0 $\pm$ 0.7 & OOM & OOM & OOT & 0.2 $\pm$ 0.1 & OOT & \textbf{95.8} $\pm$ 0.4 & 95.6 $\pm$ 0.4 \\
				& $p=0.10$ & 0.0 $\pm$ 0.1 & OOM & 21.8 $\pm$ 0.8 & OOM & 10.1 $\pm$ 2.3 & OOM & OOM & OOT & 0.1 $\pm$ 0.1 & OOT & \textbf{92.7} $\pm$ 0.4 & 92.3 $\pm$ 0.3 \\
				& $p=0.15$ & 0.0 $\pm$ 0.0 & OOM & 14.8 $\pm$ 1.1 & OOM & 3.0 $\pm$ 1.4 & OOM & OOM & OOT & 0.1 $\pm$ 0.2 & OOT & \textbf{89.4} $\pm$ 0.6 & 88.1 $\pm$ 1.2 \\
				& $p=0.20$ & 0.0 $\pm$ 0.0 & OOM & 11.2 $\pm$ 0.6 & OOM & 3.7 $\pm$ 2.0 & OOM & OOM & OOT & 0.1 $\pm$ 0.2 & OOT & \textbf{85.9} $\pm$ 0.7 & 84.8 $\pm$ 1.5 \\
				& $p=0.25$ & 0.1 $\pm$ 0.1 & OOM & 8.9 $\pm$ 0.8 & OOM & 2.6 $\pm$ 1.2 & OOM & OOM & OOT & 0.1 $\pm$ 0.2 & OOT & \textbf{82.7} $\pm$ 0.7 & 81.5 $\pm$ 1.5 \\
				\bottomrule
			\end{tabular}
		}
		\caption{Results for the subgraph-sampling benchmark. Values for $p=0$ are the same as in \Cref{tab:accuracy-summary}.}
		\label{tab:accuracy-summary_subgraph}
	\end{subtable}\vspace{-2mm}
\end{table*}

\smallskip
\noindent
\underline{\textbf{Baselines:}}
We include recent graph-alignment methods and evaluate SGWL~\cite{xu2019scalable}, PARROT~\cite{zeng2023parrot}, and FUGAL~\cite{bommakanti2024fugal} on the clique and bipartite representations, denoted by cli. and bip., respectively; e.g., SGWL cli. denotes applying SGWL to the clique representation.
For bipartite representations, we run the aligner on
the expanded node set but evaluate accuracy only on the original vertices.
We use the public implementations \footnote{SGWL and PARROT:
	\url{https://github.com/constantinosskitsas/Framework_GraphAlignment};
	FUGAL: \url{https://github.com/idea-iitd/Fugal}.}
and the suggested default parameters.
Moreover, we include BIGALIGN~\cite{koutra2013big}, designed specifically for
bipartite graph alignment; we implement its efficient \textsc{BIGALIGN-Skip}
variant in Python with the default parameters from~\cite{koutra2013big}.
We also evaluate full TAME~\cite{mohammadi2016triangular}, a triangle-tensor
graph-alignment method, on the clique representation. 
Finally, we compare against native hypergraph baselines: ELRUHNA~\cite{ibrahim2025elruhna}
is a recent unsupervised hypergraph-alignment method based on the bipartite
incidence representation.
Additionally, we include HCN+CONE, which encodes each hypergraph independently
with a two-layer HypergraphConv featureless autoencoder~\cite{bai2021hypergraph}
(embedding dim.\ 64, hidden dim.\ 128, lr 0.01, 512 epochs) and aligns the
embeddings with the CONE transformation~\cite{chen2020cone} using its default
parameters.\looseness=-1

We attempted to evaluate HyperAlign~\cite{do2024unsupervised} using the authors' public implementation, but could not reproduce reliable results, including on the authors' own experimental setting. This is consistent with direct communication with the authors, the repository's public reproducibility note, and ELRUHNA~\cite{ibrahim2025elruhna}, which excludes HyperAlign for the same reason. We therefore exclude HyperAlign.\footnote{\url{https://github.com/manhtuando97/HyperAlign}}

\smallskip
\noindent
\underline{\textbf{Experimental setup:}}
Our algorithm \hypergwl is implemented in Python 3.9 and PyTorch 2.5.1.
All experiments were run on a computer cluster. Each experiment for all algorithms ran exclusively on a node with an 
Intel(R)\,Xeon(R)\,Gold\,6130
CPU\,@\,2.10\,GHz, 384\,GB of RAM, and an NVIDIA\,A100\,GPU. 
We used a time limit of two hours.
Our code and the datasets are anonymously available at \textcolor{black}{\url{https://gitlab.com/hygr/falcon}}.

We evaluate two configurations of our framework: \hypergwlc uses the cumulative aggregation mode and \hypergwlnc uses the non-cumulative aggregation mode; both share the same solver, hyperparameters, and filtration scoring.
We use $\xi=32$ buckets, $\beta=0.1$, $K=200$ outer and $10$ inner iterations of the entropic GW solver, unless stated otherwise. These parameters were chosen via independent held-out runs.

\subsection{Results}

\noindent\textbf{RQ1: Accuracy and robustness.}
\Cref{tab:accuracy-summary,tab:accuracy-summary_subgraph}
report alignment accuracy under increasing structural perturbation. Across both benchmarks, \hypergwl matches or outperforms all baselines on
\emph{Email}, \emph{House}, and \emph{Dawn} at every nonzero noise level
(and is within one point on \emph{Email} at $p=0$), and remains competitive
on \emph{NDC}. The main pattern is robustness: while several
baselines either fail to complete on larger hypergraphs or collapse to
near-random accuracy under perturbation, \hypergwl degrades more
gradually.
At $p=0$, remaining errors arise from structural symmetries, where multiple valid structure-preserving mappings are counted against the planted permutation.\looseness=-1

TAME completes only on \emph{NDC} and at $p=0.25$ it
reaches $4.4\%$ under incidence noise and $13.2\%$ under subgraph sampling,
compared with $30.6\%$ and $38.4\%$ for \hypergwlc. This is because
TAME optimizes triangles in the clique-expanded graph; in this representation,
many triangles are induced mechanically by large hyperedges rather than being
independently observed higher-order relations.

The advantage of \hypergwl is most pronounced on large hypergraphs and at high noise. Under
incidence noise at $p=0.25$, the best \hypergwl variant retains about $82\%$
accuracy on \emph{Email}, while SGWL on the clique representation drops to
$0.2\%$ and ELRUHNA reaches $37.4\%$. On \emph{House}, no graph-based baseline
exceeds $2\%$ accuracy, whereas both \hypergwl variants remain above $60\%$ at
$p=0.25$. On \emph{Dawn}, all non-hypergraph baselines fail or are essentially
at zero accuracy, while \hypergwlnc still recovers $40.2\%$ of the
correspondences. Under subgraph sampling, \hypergwl achieves perfect or
near-perfect recovery on \emph{House}, large gains on \emph{Dawn}, and small
but consistent improvements on \emph{Email}. Thus, the benefit is not tied to a
single perturbation model, but appears under both incidence corruption and
partial hyperedge observation.
\emph{NDC} is the main exception and clarifies the limits of the approach. It is
smaller and sparser than the other datasets, so each filtration bucket contains
less redundant co-occurrence evidence. In this regime, separating the
hypergraph into many scale-specific views can produce weak individual signals,
and incidence-preserving methods such as ELRUHNA remain competitive at low
noise. As perturbation increases, however, \hypergwlc becomes strongest under
incidence noise and remains best throughout subgraph sampling, suggesting that
cumulative aggregation is preferable when structural evidence is sparse and
needs reinforcement across levels.

The comparison of \hypergwlc and \hypergwlnc gives a practical guideline. \hypergwlnc suits dense hypergraphs, where scale-specific views avoid propagating corrupted memberships to later levels, while \hypergwlc suits sparse evidence or hyperedge removal, where repeated co-occurrence across levels compensates for missing observations. 

\begin{table}[t]
	\centering
	\caption{Ablation study of \hypergwl's multi-scale representation.
		Each entry reports mean accuracy (\%) $\pm$ standard deviation over
		noise levels $p$ and ten trials per
		level. ($\dagger$~uses the ground-truth mapping to select the best
		single bucket.)}
	\label{tab:ablation}
	\begin{subtable}[t]{1\linewidth}
		\centering
		\resizebox{1\linewidth}{!}{%
			\begin{tabular}{lcccc}
				\toprule
				\textbf{Variant} & \emph{NDC} & \emph{Email} & \emph{House} & \emph{Dawn} \\
				\midrule
				First bucket only              & $ 5.1 \pm  1.4$ & $ 9.4 \pm  2.9$ & $ 6.7 \pm  4.5$ & $20.8 \pm  9.3$ \\
				Middle bucket only             & $ 5.3 \pm  2.8$ & $21.2 \pm 13.7$ & $52.9 \pm 32.5$ & $21.9 \pm 12.2$ \\
				Final bucket only              & $ 8.3 \pm  1.4$ & $32.6 \pm  3.2$ & $ 9.3 \pm  7.5$ & $13.2 \pm  5.3$ \\
				Oracle single bucket$^\dagger$ & $ 8.8 \pm  1.4$ & $33.4 \pm 12.9$ & $56.3 \pm 30.1$ & $43.1 \pm 16.3$ \\
				Pooled levels, single cost      & $29.2 \pm 12.4$ & $78.7 \pm  7.1$ & $ 3.3 \pm  4.4$ & $11.5 \pm 16.4$ \\
				\midrule
				\hypergwlc                     & $\mathbf{41.9 \pm  9.7}$ & $87.4 \pm  4.5$ & $83.7 \pm 14.9$ & $40.4 \pm 15.6$ \\
				\hypergwlnc                    & $23.8 \pm 11.9$ & $\mathbf{88.9 \pm  4.9}$ & $\mathbf{86.3 \pm 12.8}$ & $\mathbf{55.3 \pm 12.5}$ \\
				\bottomrule
		\end{tabular}}
		\caption{Incidence-noise benchmark.}
		\label{tab:ablation_incidence}
	\end{subtable}
	
	\begin{subtable}[t]{1\linewidth}
		\centering
		\resizebox{1\linewidth}{!}{%
			\begin{tabular}{lcccc}
				\toprule
				\textbf{Variant} & \emph{NDC} & \emph{Email} & \emph{House} & \emph{Dawn} \\
				\midrule
				First bucket only              & $ 3.9 \pm  1.3$ & $ 8.4 \pm  3.5$ & $18.1 \pm  8.1$ & $24.9 \pm 15.1$ \\
				Middle bucket only             & $ 4.7 \pm  2.4$ & $27.6 \pm 11.3$ & $89.6 \pm  9.1$ & $30.4 \pm  3.7$ \\
				Final bucket only              & $ 7.0 \pm  2.1$ & $31.4 \pm  4.0$ & $60.3 \pm  9.9$ & $17.1 \pm  1.6$ \\
				Oracle single bucket$^\dagger$ & $ 7.5 \pm  2.3$ & $43.6 \pm 14.2$ & $88.7 \pm 17.8$ & $48.2 \pm 19.6$ \\
				Pooled levels, single cost      & $46.2 \pm  7.4$ & $95.4 \pm  2.0$ & $85.9 \pm  6.0$ & $88.9 \pm  5.2$ \\
				\midrule
				\hypergwlc                     & $\mathbf{48.6 \pm  8.5}$ & $\mathbf{95.7 \pm  1.8}$ & $\mathbf{100.0 \pm 0.0}$ & $\mathbf{89.3 \pm 5.2}$ \\
				\hypergwlnc                    & $30.4 \pm 11.9$ & $93.8 \pm  3.2$ & $99.9 \pm  0.1$ & $88.5 \pm 5.7$ \\
				\bottomrule
		\end{tabular}}
		\caption{Subgraph-sampling benchmark.}
		\label{tab:ablation_subgraph}
	\end{subtable}\vspace{-3mm}
\end{table}

\smallskip
\noindent
\textbf{RQ2: Contribution of the multi-scale representation.}
We first test whether \hypergwl benefits from retaining filtration levels as
separate structural views. \Cref{tab:ablation} compares single-bucket,
pooled, and multi-scale variants. All variants use the same filtration score,
GW solver, and LAP decoding step; they differ only in how the
filtration-induced cost matrices are used.

The first, middle, and final variants run GW on one non-cumulative bucket only
(bucket $1$, bucket $\lceil \xi/2\rceil$, or bucket $\xi$). The oracle single
bucket chooses the best-performing bucket for each instance using the
ground-truth mapping, and is therefore not available in the unsupervised
setting. The pooled-level variant aggregates all bucket-specific cost matrices
into one collapsed cost matrix before GW: it uses information from all buckets,
but removes the scale separation. In contrast, \hypergwlc and \hypergwlnc keep
the filtration levels as separate structural channels and optimize one shared
coupling across them.
The results show that no single filtration level is reliable. The first bucket
is usually too sparse, while the middle and final buckets can be effective on
some datasets but fail on others. Even the oracle single-bucket baseline is
usually below the \hypergwl variants, showing that the benefit is not merely due
to selecting a good scale. Rather, \hypergwl gains from requiring one coupling to
explain several scale-separated views simultaneously. Pooled levels are often
stronger than individual buckets, especially under subgraph sampling, but they
can still fail when collapsing the filtration destroys scale-specific evidence,
as seen on House and Dawn under incidence noise. 
The table supports
the  design choice of \hypergwl: preserving multiple filtration-induced
views and aligning them jointly is more robust than using any single or
collapsed representation.\looseness=-1

\begin{figure*}[htbp]
	\centering
	\includegraphics[width=0.5\textwidth]{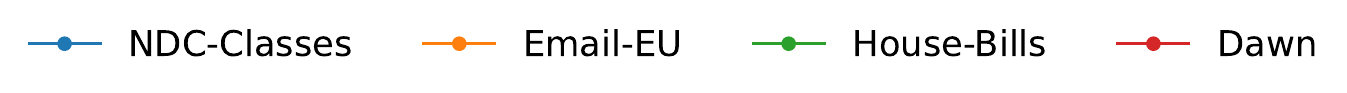}\\
	\begin{subfigure}[b]{0.24\textwidth}
		\centering
		\includegraphics[width=\textwidth]{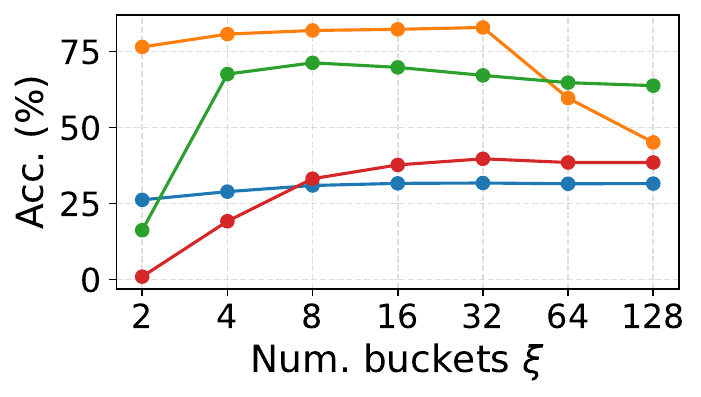}
		\caption{Incidence noise benchmark}
		\label{fig:sub1}
	\end{subfigure}
	\hfill %
	\begin{subfigure}[b]{0.24\textwidth}
		\centering
		\includegraphics[width=\textwidth]{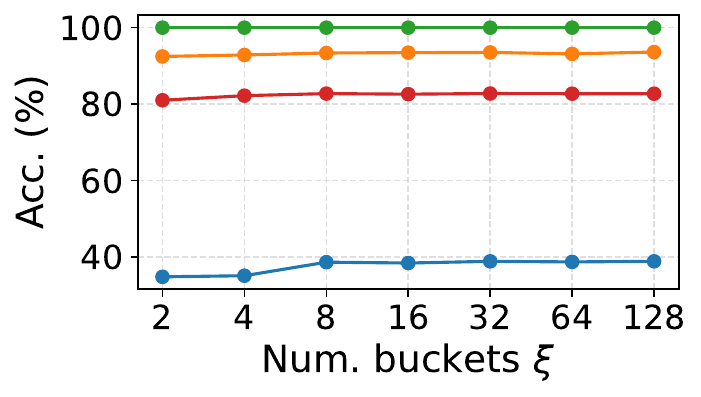}
		\caption{Subgraph-sampling benchm.}
		\label{fig:sub2}
	\end{subfigure}
	\hfill
	\begin{subfigure}[b]{0.24\textwidth}
		\centering
		\includegraphics[width=\textwidth]{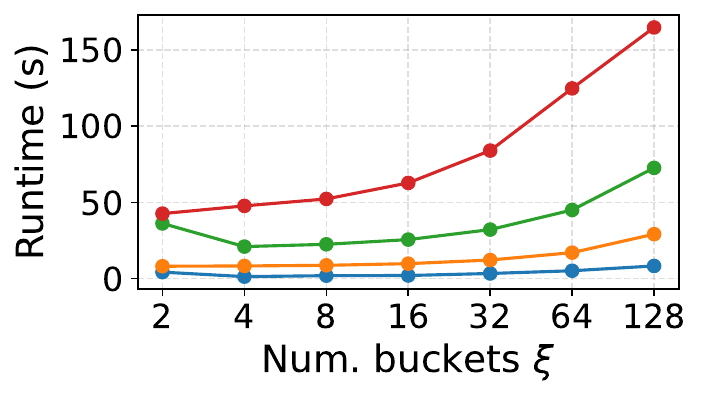}
		\caption{Incidence noise benchmark}
		\label{fig:sub3}
	\end{subfigure}
	\hfill
	\begin{subfigure}[b]{0.24\textwidth}
		\centering
		\includegraphics[width=\textwidth]{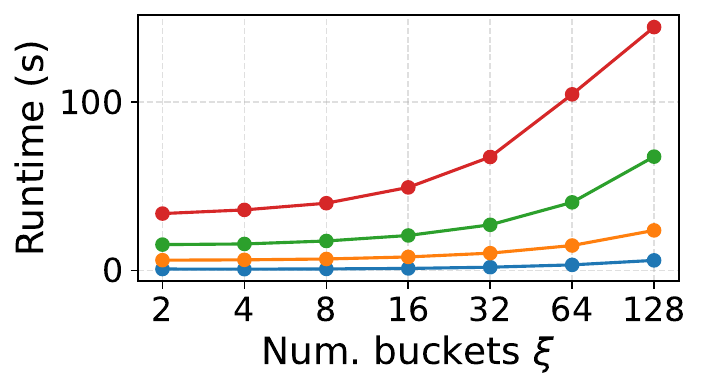}
		\caption{Subgraph-sampling benchm.}
		\label{fig:sub4}
	\end{subfigure}
	\caption{Impact of the number of buckets $\xi$ on accuracy and running time ($p=0.25$, ten independent runs).}
	\label{fig:ab}
\end{figure*}

\smallskip
\noindent
\textbf{Sensitivity to the number of buckets.}
We next study the sensitivity of \hypergwl to the filtration resolution $\xi$.
For this diagnostic experiment, we vary
$\xi \in \{2,4,8,16,32,64,128\}$ and report results at the highest noise level,
$p=0.25$. To isolate the effect of $\xi$ from the choice of aggregation mode,
we use, for each dataset and benchmark, the stronger \hypergwl variant
(\hypergwlc or \hypergwlnc) under the default configuration.
\Cref{fig:sub1,fig:sub2} show the impact on accuracy. Under incidence noise,
very coarse filtrations are often insufficient, and accuracy improves as the
filtration is refined, especially on \emph{House} and \emph{Dawn}. The gains
then largely saturate around $\xi \in \{16,32\}$. Increasing $\xi$ further adds
little and can hurt performance, most clearly on \emph{Email}, where very fine
buckets appear to introduce noisy, weakly populated levels. Under subgraph
sampling, the sensitivity is weaker: accuracy is already high for small
$\xi$ and remains comparatively stable as the number of buckets increases.
Together with the corresponding runtime curves in \Cref{fig:sub3,fig:sub4},
these results support using a moderate default value, $\xi=32$, which lies near
the empirical accuracy plateau while avoiding the additional cost and possible
instability of overly fine filtrations.

\begin{table*}[t]
	\centering
	\caption{Ablation of filtration score, scale weighting, dissimilarity function, and aggregation
		mode. Mean accuracy (\%) $\pm$ standard deviation over noise levels $p \in \{0.05, 0.10, 0.15, 0.20, 0.25\}$.
		Bold = best per column.}
	\label{tab:ablation_design_combined}
	\setlength{\tabcolsep}{3pt}
	\renewcommand{\arraystretch}{1.1}
	\resizebox{1\linewidth}{!}{%
		\begin{tabular}{lcccccccccccc}
			\toprule
			& & & & &
			\multicolumn{4}{c}{\textbf{Incidence noise}} &
			\multicolumn{4}{c}{\textbf{Subgraph sampling}} \\
			\cmidrule(lr){6-9}\cmidrule(lr){10-13}
			& \textbf{Filtration} & \textbf{Weighting} & \textbf{Dissimilarity} & \textbf{Cumu.}
			& \emph{NDC} & \emph{Email} & \emph{House} & \emph{Dawn}
			& \emph{NDC} & \emph{Email} & \emph{House} & \emph{Dawn} \\
			\midrule
			& Size       & balanced & binary  & \cmark
			& $38.7 \pm 10.7$ & $76.0 \pm  8.4$ & $69.4 \pm 28.2$ & $10.8 \pm 13.8$
			& $48.4 \pm  8.1$ & $94.6 \pm  4.6$ & $99.9 \pm  0.0$ & $87.7 \pm  5.8$ \\
			& Size       & balanced & binary  & \xmark
			& $17.6 \pm 12.9$ & $73.9 \pm  9.6$ & $83.5 \pm 14.2$ & $ 6.3 \pm  5.6$
			& $46.8 \pm 12.6$ & $93.5 \pm  8.1$ & $\mathbf{100.0 \pm 0.0}$ & $88.4 \pm  5.5$ \\
			\midrule
			& Deg.-aware & uniform  & binary  & \cmark
			& $41.1 \pm  8.5$ & $84.7 \pm  4.7$ & $82.2 \pm 14.0$ & $34.4 \pm 17.8$
			& $48.4 \pm  7.6$ & $95.5 \pm  1.9$ & $99.9 \pm  0.0$ & $89.0 \pm  4.8$ \\
			& Deg.-aware & uniform  & binary  & \xmark
			& $24.3 \pm 11.4$ & $82.7 \pm  8.0$ & $85.7 \pm 12.1$ & $52.2 \pm 11.4$
			& $30.4 \pm 11.5$ & $93.9 \pm  2.9$ & $99.8 \pm  0.1$ & $86.9 \pm  5.9$ \\
			\midrule
			& Deg.-aware & balanced & Jaccard & \cmark
			& $21.9 \pm  8.1$ & $29.4 \pm  7.3$ & $70.5 \pm  6.6$ & $14.3 \pm  2.0$
			& $30.8 \pm  8.7$ & $25.8 \pm  8.4$ & $77.3 \pm  7.3$ & $14.6 \pm  2.3$ \\
			& Deg.-aware & balanced & Jaccard & \xmark
			& $21.8 \pm  9.4$ & $48.8 \pm 12.0$ & $78.2 \pm 10.5$ & $23.6 \pm  3.6$
			& $27.2 \pm  9.3$ & $63.9 \pm 10.7$ & $93.6 \pm  2.8$ & $31.3 \pm  6.8$ \\
			\midrule
			\hypergwlc  & Deg.-aware & balanced & binary  & \cmark
			& $\mathbf{41.9 \pm 9.7}$ & $87.4 \pm 4.5$ & $83.7 \pm 14.9$ & $40.4 \pm 15.6$
			& $\textbf{48.6} \pm 8.5$ & $\textbf{95.7} \pm 1.8$ & $\mathbf{100.0 \pm 0.0}$ & $\textbf{89.3} \pm 5.2$ \\
			\hypergwlnc & Deg.-aware & balanced & binary  & \xmark
			&$23.8 \pm 11.9$ & $\mathbf{88.9 \pm 4.9}$ & $\mathbf{86.3 \pm 12.8}$ & $\mathbf{55.3 \pm 12.5}$
			& $30.4 \pm 11.9$ & $93.8 \pm 3.2$ & $99.9 \pm 0.1$ & $88.5 \pm 5.7$ \\
			\bottomrule
	\end{tabular}}
\end{table*}

\smallskip
\noindent
\textbf{Impact of the filtration score, weighting, and dissimilarity function:}
To assess each design choice, we compare against the most natural simpler alternatives: hyperedge size as a filtration score (instead of degree-aware ordering  (Eq.\,\ref{eq:deg})), uniform scale weighting (instead of balanced edge-count weighting (Eq.\,\ref{eq:weight})), and Jaccard distance as dissimilarity function (instead of binary overlap (Eq.\,\ref{eq:overlap})). Table~\ref{tab:ablation_design_combined} reports the results. The choice of dissimilarity function has the largest impact: replacing binary overlap with Jaccard causes substantial accuracy drops across both benchmarks, most severely on Dawn (55.3\%~$\to$~23.6\% under incidence noise) and Email (89.0\%~$\to$~48.8\%), confirming that ignoring co-occurrence multiplicity is beneficial under structural noise. The filtration score matters most on Dawn, where switching from degree-aware to size-based ordering collapses accuracy from 55.3\% to 6.3\% under incidence noise, while the effect is modest on House where both scores perform similarly. Scale weighting has the smallest impact overall: balanced edge-count weighting matches or slightly outperforms uniform weighting in most configurations, with differences rarely exceeding 2--3 percentage points. Across both benchmarks, \hypergwl matches or exceeds all ablated variants on all four datasets.

\begin{table}
	\centering
	\caption{Runtime in seconds (mean $\pm$ std.~over
		successful runs). OOT---out of time, OOM---out of memory.}
	\label{tab:runtime}
	\renewcommand{\arraystretch}{1.0}		\resizebox{1\linewidth}{!}{%
		\begin{tabular}{lr@{ $\pm$ }lr@{ $\pm$ }lr@{ $\pm$ }lr@{ $\pm$ }l}
			\toprule
			\textbf{Algorithm} & \multicolumn{2}{c}{\emph{NDC}} & \multicolumn{2}{c}{\emph{Email}} & \multicolumn{2}{c}{\emph{House}} & \multicolumn{2}{c}{\emph{Dawn}} \\
			\midrule
			SGWL cli.     & 4.6 & 0.6 & 41.1 & 21.1 & 67.4 & 11.4 & 69.3 & 1.9 \\
			SGWL bip.    & 47.0 & 0.2 & \multicolumn{2}{c}{OOT} & \multicolumn{2}{c}{OOT} & \multicolumn{2}{c}{OOM} \\
			PARROT cli.   & 1.7 & 0.4 & 22.7 & 7.4 & 3550.8 & 0.0 & 181.1 & 0.0 \\
			PARROT bip.  & \textbf{0.5} & 0.0 & 19.1 & 0.2 & 758.6 & 0.0 & 1064.7 & 0.0 \\
			FUGAL cli.    & 16.2 & 0.6 & 54.3 & 10.7 & 3882.7 & 0.0 & 338.1 & 117.0 \\
			FUGAL bip.   & 56.2 & 0.7 & \multicolumn{2}{c}{OOT} & \multicolumn{2}{c}{OOT} & \multicolumn{2}{c}{OOM} \\
			BIGALIGN     & 6.0 & 0.2 & 1941.1 & 11.4 & \multicolumn{2}{c}{OOT} & \multicolumn{2}{c}{OOM} \\
			TAME		 &123.9 & 31.0 & \multicolumn{2}{c}{OOT} & \multicolumn{2}{c}{OOT} & \multicolumn{2}{c}{OOT} \\
			HCN+CONE     & 4.2 & 0.0 & \textbf{9.8} & 0.3 & 61.8 & 2.0 & \textbf{48.9} & 1.5 \\
			ELRUHNA      & 5.2 & 0.4 & 2985.7 & 100.6 & \multicolumn{2}{c}{OOT} & \multicolumn{2}{c}{OOT} \\
			\hypergwlc   & 1.4 & 0.1 & 14.1 & 3.2 & 33.5 & 0.3 & 80.4 & 5.3 \\
			\hypergwlnc  & 1.4 & 0.1 & 13.4 & 1.0 & \textbf{33.4} & 0.9 & 80.1 & 5.4 \\
			\bottomrule
		\end{tabular}
	}%
\end{table}

\begin{figure}
	\centering
	\includegraphics[width=0.68\linewidth]{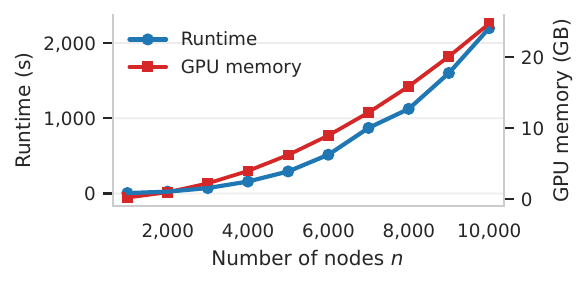}
	\caption{Runtime and memory usage of \hypergwlc with increasing nodes $n$. Number of hyperedges fixed at $m=10^5$.}
	\label{fig:scaling}\vspace{-2mm}
\end{figure}

\smallskip
\noindent
\textbf{RQ3: Efficiency.}
\Cref{tab:runtime} reports mean running time over noise levels and trials.
Although \hypergwl is not always the fastest method on the smallest dataset, it is
the only method that both completes within the time limit and remains accurate
across all datasets and both perturbation benchmarks. The clearest scalability
issue is the bipartite representation: by expanding the alignment problem from
$|V|$ to $|V|+|E|$ nodes, SGWL bip., FUGAL bip., BIGALIGN, and ELRUHNA become
infeasible on the larger datasets. Clique-based methods avoid this node blow-up, but can still be slow on dense clique
expansions: PARROT and FUGAL require several thousand seconds on \emph{House},
and TAME completes only on \emph{NDC} ($123.9$s) while timing out elsewhere
because its triangle-tensor updates scale with the number of triangles induced
by the clique graph. In contrast, \hypergwl is fastest on \emph{House} and
remains practical on \emph{Email} and \emph{Dawn}.

\Cref{fig:scaling} complements these comparisons with a controlled scaling
experiment for \hypergwlc using synthetic hypergraphs. With the number of hyperedges fixed at $m=10^5$ and random hyperedge sizes between 2 and 8 nodes,
runtime and GPU memory increase smoothly with the number of nodes, reaching the
largest tested setting without exhausting memory. 
This does not remove the intrinsic scalability ceiling of alignment: \hypergwl
still solves a dense GW problem (a relaxation of the quadratic assignment problem) on the original node set, whose
dominant cost grows cubically in $n$ and instances beyond the range shown in \Cref{fig:scaling} become increasingly demanding.

\section{Conclusion}
We introduced \hypergwl{}, a fully unsupervised framework for hypergraph alignment based on filtration-induced co-occurrence views and a shared multi-scale GW objective.
By optimizing one transport plan across all filtration levels, \hypergwl{} produces a globally consistent node correspondence while preserving structural evidence across scales.
Our experiments showed that this multi-scale representation improves robustness to structural perturbations and performs competitively against strong graph- and hypergraph-alignment baselines.

\medskip
\noindent
\textbf{Acknowledgments.}
This research was supported by 
the ERC Advanced Grant {\small REBOUND} (834862),
the Swedish Research Council project {\small ExCLUS} (2024-05603),
and the Wallenberg AI, Autonomous Systems and Software Program ({\small WASP}) funded by the Knut and Alice Wallenberg Foundation.

\medskip
\noindent
\textbf{Acknowledgement of GenAI Use.}
Large language models (LLMs) were used during the preparation of this paper to assist with writing, code development, and debugging.

\balance

\end{document}